\documentclass[conference]{IEEEtran}
\IEEEoverridecommandlockouts
\usepackage{CJKutf8}
\usepackage{cite}
\usepackage{amsmath,amssymb,amsfonts}
\usepackage{algorithmic}
\usepackage{graphicx}
\usepackage{textcomp}
\usepackage{multicol}
\usepackage{multirow}
\usepackage{xcolor}
\usepackage{makecell}
\usepackage{changepage}
\usepackage{threeparttable}
\usepackage{hhline}
\usepackage{subfigure}
\usepackage[ruled,vlined,linesnumbered]{algorithm2e}
\def\BibTeX{{\rm B\kern-.05em{\sc i\kern-.025em b}\kern-.08em
		T\kern-.1667em\lower.7ex\hbox{E}\kern-.125emX}}

\makeatletter
\newcommand{\linebreakand}{%
\end{@IEEEauthorhalign}
\hfill\mbox{}\par
\mbox{}\hfill\begin{@IEEEauthorhalign}
}
\makeatother

\begin{document}
\begin{CJK}{UTF8}{gbsn}

	\title{An Efficient Reconstructed Differential Evolution Variant by Some of the Current State-of-the-art Strategies for Solving Single Objective Bound Constrained Problems}
	
	\author{
		\IEEEauthorblockN{1\textsuperscript{st} Sichen Tao} \IEEEauthorblockA{\textit{Faculty of Engineering} \\ \textit{University of Toyama} \\ Toyama, Japan \\ taosc73@hotmail.com \\ d2278002@ems.u-toyama.ac.jp}%
		\and
		\IEEEauthorblockN{2\textsuperscript{nd} Ruihan Zhao} \IEEEauthorblockA{\textit{School of Mechanical Engineering} \\ \textit{Tongji University} \\ Shanghai, China \\ 2110415@tongji.edu.cn}%
		\and
		\IEEEauthorblockN{3\textsuperscript{rd} Kaiyu Wang} \IEEEauthorblockA{\textit{Faculty of Engineering} \\ \textit{University of Toyama} \\ Toyama, Japan \\ greskofairy@gmail.com}%
            \and
		\IEEEauthorblockN{4\textsuperscript{th} Shangce Gao} \IEEEauthorblockA{\textit{Faculty of Engineering} \\ \textit{University of Toyama} \\ Toyama, Japan \\ gaosc@eng.u-toyama.ac.jp}
	}
	
	\maketitle

\begin{abstract}
Complex single-objective bounded problems are often difficult to solve. In evolutionary computation methods, since the proposal of differential evolution algorithm in 1997, it has been widely studied and developed due to its simplicity and efficiency. These developments include various adaptive strategies, operator improvements, and the introduction of other search methods. After 2014, research based on LSHADE has also been widely studied by researchers. However, although recently proposed improvement strategies have shown superiority over their previous generation's first performance, adding all new strategies may not necessarily bring the strongest performance. Therefore, we recombine some effective advances based on advanced differential evolution variants in recent years and finally determine an effective combination scheme to further promote the performance of differential evolution. In this paper, we propose a strategy recombination and reconstruction differential evolution algorithm called reconstructed differential evolution (RDE) to solve single-objective bounded optimization problems. Based on the benchmark suite of the 2024 IEEE Congress on Evolutionary Computation (CEC2024), we tested RDE and several other advanced differential evolution variants. The experimental results show that RDE has superior performance in solving complex optimization problems.
\end{abstract}


\section{Introduction}
Optimization has become a fundamental tool in applied mathematics, engineering, economics, medical science and other scientific fields, including energy systems, bioinformatics, finance, manufacturing and production, transportation and logistics, telecommunications networks etc\cite{gao2016ant}\cite{gao2021state}. Real-world optimization problems are usually related to one or more of the following: nonlinear complex constraints, multimodal, multidimensional, non-differentiable, and noisy search spaces. Traditional deterministic methods often perform poorly in complex optimization problems. In contrast, evolutionary computation methods have been discovered and considered as an excellent alternative to traditional optimization techniques in the past 40 years of research\cite{8409490}. They can avoid local optima as much as possible within reasonable computing time and find at least sufficiently good solutions\cite{8937719}\cite{wang2023spherical}. Additionally, evolutionary algorithms are simple and easy to use. They are usually based on some simple biological or object concepts without requiring differentiation of problem functions. They have certain generality that can be used to solve various types of optimization problems.

Several classic optimization methods based on evolutionary thinking have been proposed, such as genetic algorithm\cite{sohail2023genetic}, particle swarm optimization\cite{yarat2021comparative}, and differential evolution (DE)\cite{ahmad2022differential}. Their various advanced variants have also been fully developed, tested, and researched over the past 20 years. In the IEEE Congress on Evolutionary Computation (IEEE CEC) in recent decades, various types of optimization problems have been widely tested and competed. Among them, since SHADE was proposed in 2013\cite{tanabe2013success}, new variants of differential evolution algorithms have consistently ranked high in annual competitions. The encoding of these algorithms is not very complex but they exhibit excellent performance on relatively complex space optimization problems and continue to achieve performance improvements. A notable improvement path has emerged from this research. In 2014, LSHADE\cite{tanabe2014improving} improved the upper limit performance of DE variants in the CEC2014 standard test set by introducing linear population changes. In 2018, LSHADE-RSP\cite{stanovov2018lshade} achieved second place in the competition by introducing a rank-based selection pressure strategy. In 2021, an improved LSHADE-RSP (iLSHADE-RSP) using Cauchy perturbation search was shown to further improve the upper limit performance of this series of DEs \cite{choi2021improved}. Additionally, in 2019 an enhanced mutation strategy called EB mutation strategy \cite{mohamed2019novel} was reported to effectively improve the performance of LSHADE and its previous generations' algorithms as well.

In this study, based on recent research that has brought effective performance improvements to DE advanced variants, we construct and propose a new DE variant called Reconstructed Differential Evolution (RDE). We studied multiple parameter adaptive strategies and developed the RSP strategy. In addition, the EB mutation strategy was also introduced to participate in the optimization process with LSHADE series mutation strategies, and adaptive control of dual-operator population resources was achieved through an evaluation method based on fitness progress. We tested RDE on the benchmark suite of the 2024 IEEE Congress on Evolutionary Computation (CEC2024) and compared it with other advanced evolutionary algorithms. The experimental results show that RDE has excellent performance and development potential in solving single-objective bounded optimization problems.

\section{Reconstruct Differential Evolution} 
\subsection{The Basic DE of RDE}
In this section, we elaborate on one of the classic variants of DE and use it as the basis for our research in this paper. In the following sections, we will gradually develop based on this basic DE.

Like other evolutionary algorithms, DE typically generates an initial population of $N$ individuals based on the following formula.
\begin{equation} \label{eq1}
    x_{i,j} = rand(x_{L,j},x_{U,j})
\end{equation}
where the function $rand(a,b)$ returns a random number within the range of $[a,b]$. $x_{L,j}$ and $x_{U,j}$ respectively represent the lower bound and upper bound of dimension $j$.

The core of differential evolution is the mutation operator. Since the original differential evolution (DE) was proposed in 1997 \cite{storn1997differential}, various improved mutation operators have been introduced, including DE/rand/1, DE/rand/2, DE/best/1, DE/best/2, and DE/current-to-best/1. The most famous one is the DE/current-to-pbest/1 mutation strategy introduced by JADE in 2009 \cite{zhang2009jade}. This strategy is also the most commonly used efficient strategy for LSHADE and its descendant variants. Therefore, we use it as the basis of our research in this paper. Its expression is as follows:
\begin{equation} \label{eq2}
    v_{i,j}^{(k+1)}=x_{i,j}^{(k)}+F\left(x_{p,j}^{(k)} - x_{i,j}^{(k)}\right)+F\left(x_{r1,j}^{(k)}-x_{r2,j}^{(k)}\right)
\end{equation}
where $k$ represents the number of iterations. $x_{i,j}$ refers to the value of the $j$-th dimension of the $i$-th individual. $r1$ and $r2$ are two different random indices used for randomly selecting individuals from the population. $p$ is a randomly selected index value from the top $100p\%$ individuals in the population sorted by fitness values, with a range of $(0,1]$. $F$ is a scale factor used to control the size of differential vectors, with a range of $[0,1]$.

Another major part of LSHADE is crossover, which is controlled by a crossover probability $Cr$ ranging from 0 to 1. For each dimension of every new trial solution generated after mutation for each individual in the population, there is a probability of $(1-Cr)$ that it will be replaced by the value of the same dimension from its parent. This inheritance comes from earlier genetic algorithms in evolutionary computation \cite{holland1992genetic}. The expression is as follows:
\begin{equation} \label{eq3}
	u_{i,j}^{(k)}\;=\;\left\{\begin{array}{l}v_{i,j}^{(k)},\;\;\;j\;=\;j_{rand}\;\;{\rm or}\;\;rand\;<\;Cr\\x_{i,j}^{(k)},\;\;\;{\rm otherwise}\end{array}\right.
\end{equation}
where $u_{i,j}$ is the new individual generated by executing crossover strategy between $v_{i,j}$ and $x_{i,j}$. $j_{rand}$ represents a random integer within the range of $[0,D]$ (where $D$ is the dimension of the optimization problem). The purpose of $j_{rand}$ is to ensure that at least one dimension adopts the value generated by mutation strategy.

Finally, there is the selection part. LSHADE compares the fitness values of offspring individuals with their corresponding parent individuals and keeps the one with better fitness value. The algorithm will perform this selection operation on each individual in the population separately. Its expression is as follows.
\begin{equation}\label{eq4} 
	x_i^{(k+1)}\;=\;\left\{\begin{array}{l}u_i^{(k)},\;\;\;f(u_i^{(k)})\;\leq\;f(x_i^{(k)})\\x_i^{(k)},\;\;\;{\rm otherwise}\end{array}\right.
\end{equation}    
where $f(u_i^{(k)})$ and $f(x_i^{(k)})$ indicates the evaluation function of the individual $u_i^{(k)}$ and $x_i^{(k)}$.

\subsection{The Reconstructured Differential Evolution}
In this section, we will combine different strategies of basic DE with various advanced DE variants proposed in recent years, and develop and propose a reconstructed DE (RDE).

\subsubsection{External Archive}
In the mutation strategy of the basic DE framework we adopt, a random index $r2$ is used to select useful individuals from a joint population consisting of the current population and an external archive. This approach further balances the algorithm's development and exploratory performance, and enhances its stability. The external archive preserves some parent individuals from history that have already been replaced in the current archive. The use of an external archive has been adopted by a series of advanced DE algorithms starting with SHADE. The size of this archive is defined as $Ar\cdot N$, where $N$ represents the number of individuals in the population.

\subsubsection{Introducing the DE/current-to-order-pbest/1 Mutation Strategy}
In a study conducted in 2019, a new mutation strategy called DE/current-to-order-pbest/1 was proposed. It sorts and recombines the first, third, and fourth terms of two differential groups in DE/current-to-pbest/1 according to their fitness values, organizes them in order of better, medium, and worse solutions, and generates applicable solutions accordingly. The expression is as follows:

\begin{equation} \label{eq5}
\begin{aligned}
v_{i,j}^{(k+1)}=x_{i,j}^{(k)}+F\left(x_{{ord\_pbest},j}^{(k)} - x_{i,j}^{(k)}\right)\\
    +F\left(x_{{ord\_median},j}^{(k)}-x_{{ord\_worst},j}^{(k)}\right)
\end{aligned}
\end{equation}

$x_{{ord\_pbest},j}^{(k)}$, $x_{{ord\_median},j}^{(k)}$ and $x_{{ord\_worst},j}^{(k)}$ represent the individuals in the corresponding positions after sorting, which correspond to better, medium and worse fitness values among the three.

\subsubsection{Hybridization of Two Mutation Strategies}
We use both effective mutation strategies mentioned above to update individuals in the population. Here, we propose a simple adaptive strategy to dynamically control the allocation ratio of population resources between the two strategies. We use $\gamma_1 \in [0,1]$ to represent the resource allocation ratio for DE/current-to-order-pbest/1 strategy, and the resource allocation ratio for DE/current-to-pbest/1 is $\gamma_2 = (1-\gamma_1)$. We use $\omega_{m1}$ and $\omega_{m2}$ to represent the average fitness improvement caused by the two mutation strategies respectively, and determine the allocation ratio of next generation based on their proportion. The expression is as follows:
\begin{equation} \label{eq6}
\omega_{m1}^{(k)} = \frac{\sum_{i=1}^{N_1} (f_{m1,i}^{(k)} - f_{m1,i}^{(k-1)})}{N_1}
\end{equation}
\begin{equation} \label{eq7}
\omega_{m2}^{(k)} = \frac{\sum_{i=1}^{N_2} (f_{m2,i}^{(k)} - f_{m2,i}^{(k-1)})}{N_2}
\end{equation}

\begin{equation} \label{eq8}
	\gamma_1^{(k+1)}\;=\;\left\{\begin{array}{l}0.5,\;\;\;\omega_{m1}^{(k)} = \omega_{m2}^{(k-1)} = 0\;or\;k = 1\\\frac{\omega_{m1}^{(k)}}{\omega_{m1}^{(k)} + \omega_{m2}^{(k-1)}},\;\;\;{\rm otherwise}\end{array}\right.
\end{equation}
where when the average fitness improvement caused by both strategies is zero, let $\gamma_1 = \gamma_2 = 0.5$. When initializing the population, $\gamma_1^{(1)} = \gamma_2^{(1)} = 0.5$.

\subsubsection{Extened Rank-based Selective Pressure Strategy}
In LSHADE-RSP\cite{stanovov2018lshade}, a fitness-based rank pressure method\cite{jebari2013selection} is proposed to correct the probability of different individuals being selected in DE, with rank indicators assigned as follows:
\begin{equation} \label{eq9}
    Rank_i = k_r \cdot (N - i) + 1
\end{equation}
where the $i$ here indicates the original rank calculated by the fitness values of different individuals in a population. $Rank_i$ indicates the corrected rank value of the individual ranked $i$th. By this equation, the scaling factor $k_r$ is responsible for the greediness of the rank selection.

Then, the probability of selecting corresponding individuals is no longer equal and is modified by the following formula:
\begin{equation} \label{eq10}
    pr_i = Rank_i / (Rank_1 + Rank_2 + ... + Rank_N)
\end{equation}
$pr_i$ represents the probability of the individual ranked $i$ being selected in each random selection of the mutation strategy.

The strategy is used in LSHADE-RSP to influence the selection probability of the last two terms in the differential term, namely $r_1$ and $r_2$ in Eq.~\ref{eq2}. However, we extend it to three random selection behaviors of two differential terms in the new variant proposed in this paper, including $p$, $r_1$, and $r_2$ in Eq.~\ref{eq2}. Correspondingly, in DE/current-to-order-pbest/1 Mutation Strategy, pre-sorting selection is also based on this Extended Rank-based Selective Pressure Strategy. Note that the selection behavior in the historical archive of LSHADE-RSP did not execute RSP. Here we also make corrections to the selection probability of historical archives using the same method.

\subsubsection{Success Historical Memory based Parameter Adaptive Strategy}
In order to adaptively adjust the values of parameters $F$ and $Cr$, the Success Historical Memory based Parameter Adaptive Strategy in jSO\cite{brest2017single} is used in RDE. Firstly, two archives with $H$ entries are used to store reference values for $F$ and $Cr$ ($H_F = H_{Cr} = H$), which are updated in real-time. The values stored at position $h$, denoted as $\mu_{F,h}$ and $\mu_{Cr,h}$ respectively ($h \in [1,H]$), are used to determine the value of $F_i$ and $Cr_i$ for each individual in the population using the following formulae during each generation:
\begin{equation}\label{eq11}
    Cr_i^{(k)} = C\!auchy\!Rand(\mu_{F,h},0.1)
\end{equation}
\begin{equation}\label{eq12}
    F_i^{(k)} = N\!ormal\!Rand(\mu_{Cr,h},0.1)
\end{equation}
\begin{equation}\label{eq13}
    h = mod(\frac{k}{H})
\end{equation}
Among them, $C\!auchy\!Rand(a,b)$ returns a random number obtained from a Cauchy distribution with median of $a$ and scale factor of $b$. $N\!ormal\!Rand(a,b)$ returns a random number obtained from a normal distribution with mean of $a$ and variance of $b$. $mod(\frac{a}{b})$ returns the remainder obtained when dividing $a$ by $b$.

Then, the used archive slot is immediately updated, expressed as follows:
\begin{equation}\label{eq14}
\mu_{F,j}^{(k)} = \frac{\sum_{n = 1}^{|S_F^{(k)}|}\omega_n^{(k)}{s_{F,n}^{(k)}}^2}{\sum_{n = 1}^{|S_F^{(k)}|}\omega_n^{(k)}s_{F,n}^{(k)}}
\end{equation}

\begin{equation}\label{eq15}
\mu_{Cr,j}^{(k)} = \frac{\sum_{n = 1}^{|S_{Cr}^{(k)}|}\omega_n^{(k)}{s_{Cr,n}^{(k)}}^2}{\sum_{n = 1}^{|S_{Cr}^{(k)}|}\omega_n^{(k)}s_{Cr,n}^{(k)}}
\end{equation}

\noindent where $s_{F,n}^{(k)}$ and $s_{Cr,n}^{(k)}$ represent the $n$th values of $F$ and $Cr$ of one successful trial in the $k$th iteration. $|S_r^{(k)}|$ and $|S_s^{(k)}|$ represent the length of success history vectors $S_r^{(k)}$ and $S_s^{(k)}$ in the $k$th iteration. $\omega_h^{(k)}$ is calculated as follows:

\begin{equation}\label{eq16}
\omega_n^{(k)} = \frac{f_n^{(k)} - f_n^{(k-1)}}{\sum_{g = 1}^{|S^{(k)}|}f_g^{(k)} - f_g^{(k-1)}}
\end{equation}

\noindent where $\omega_n$ represents the $n$th individual's fitness successful weight for evaluations in Eq.~(\ref{eq14}) and Eq.~(\ref{eq15}).

where a strategy used in iLSHADE\cite{brest2016shade}, jSO\cite{brest2017single} and LSHADE-RSP\cite{stanovov2018lshade} is introduced. The last positions of $F$ and $Cr$ in the archive are fixed during the optimization process, with their values set to 0.9 as follows:
\begin{equation}\label{eq17}
    H_{F, H} = H_{Cr, H} = 0.9
\end{equation}
In addition, some related constraints have also been adopted. In the early stages, excessively large $F$ and excessively small $Cr$ were not allowed. Their formulas are expressed as follows:
\begin{equation}\label{eq18}
	F_i^{(k)}=0.7,~~~i\!f~n\!f\!es < 0.6 \cdot max\_n\!f\!es~and~F_i^{(k)} > 0.7
\end{equation}
\begin{equation} \label{eq19}
	Cr_i^{(k)}\;=\;\left\{
        \begin{array}{l}
            0.7,\;\;\;~~~i\!f~n\!f\!es < 0.25 \cdot max\_n\!f\!es~\\
            ~~~~~~~~~~and~F_i^{(k)} < 0.7\\
            0.6,\;\;\;~~~else~i\!f~n\!f\!es < 0.5 \cdot max\_n\!f\!es~\\
            ~~~~~~~~~~and~F_i^{(k)} < 0.6
        \end{array}\right.
\end{equation}

where $n\!f\!es$ and $max\_n\!f\!es$ represent the current fitness evaluation number and the maximum fitness evaluation number.

\subsubsection{Linear $p$ Value Reduction}
After each generation of iteration $k$, the value of $p$ for "pbest" selection strategy in Eq.~(\ref{eq2}) and Eq.~(\ref{eq3}) is linearly reduced. The formula is expressed as follows:
\begin{equation}\label{eq20}
    p^{(k)} = p_{max} \cdot (1 - 0.5 \cdot \frac{n\!f\!es}{max\_n\!f\!es})
\end{equation}
where $p_{max}$ represents the initialized $p$ value. $p^{(k)}$ denotes the $p$ value for the "pbest" selection strategy at the $k$th iteration.

\subsubsection{Linear Population Size Reduction}
We introduced the linear population size reduction strategy from LSHADE.
\begin{equation}\label{eq21} 
	N^{(k+1)}\;=\;round\;((\frac{N_{min}^{(k)}\;-\;N_{max}^{(k)}}{max\_n\!f\!es})\;\cdot\;n\!f\!es\;+\;N_{max}^{(k)})
\end{equation}
where $round()$ returns one rounded integer. $N_{max}$ and $N_{min}$ denote the beginning and the ending size of the population, respectively. The value of $N_{min}$ is determined by the minimum number of population individuals that can undergo differential evolution mutation operator.

\subsubsection{Cauthy Perturbation}
A Cauthy perturbation strategy called iLSHADE-RSP, proposed in 2021, is used to replace individuals that undergo parent crossover with a probability of $p_r$. We introduce this strategy into RDE to appropriately enhance population diversity during the optimization process. The strategy is expressed as follows:
\begin{equation} \label{eq22}
	u_{i,j}^{(k)}\;=\;\left\{
    \begin{array}{l}
    v_{i,j}^{(k)},\;\;\;j\;=\;j_{rand}\;\;{or}\;\;rand\;<\;Cr_i \\
    C\!auchy\!Rand(x_{i,j}^{(k)},0.1),\;\;\;{else~i\!f~ rand < p_r}\\
    x_{i,j}^{(k)},\;\;\;{otherwise}
    \end{array}\right.
\end{equation}
where $rand$ denotes a random value from 0 to 1. $p_r$ represents a parameter to control the Cauthy perturbation strategy here.
\begin{algorithm}[!ht]
	\caption{RDE}
	\label{RDE}
	\SetAlgoLined
	\KwIn{Parameters $max\_n\!f\!es$, $D$, $N_{max} = 18D$, $H = 5$, $p_{max} = 0.25$, $Ar = 1$, $\mu_F = 0.3$, $\mu_{Cr} = 0.8$, $\gamma_1 = \gamma_2 = 0.5$, $k_r = 3$ and $p_r = 0.2$}
	\KwOut{The obtained best solution}
	\textbf{Initialization}: Randomly generate a population \{$x_1,\;x_2,\;...,\;x_{N}$\} by Eq.~(\ref{eq1}), set $n\!f\!es = 0$
	
	\While{$n\!f\!es < max\_n\!f\!es$} {
            $k$ = $k$ + 1;\\
		\For{i = 1 to $N$}{
		      Calculate the $p$ value by Eq.~(\ref{eq20});
                
                Calculate the $F_i^{(k)}$ and $Cr_i^{(k)}$ by Eq.~(\ref{eq11})-(\ref{eq13}), (\ref{eq18}) and (\ref{eq19});

                Calculate the rank-based selective pressure rand indices by Eq.~(\ref{eq9}) and (\ref{eq10});
                
			Generate the mutation trial vectors $v_{i,k}$ by Eq.~(\ref{eq2}) and (\ref{eq5});
			
			Generate the candidate solution $u_{i,j}$ uses Eq.~(\ref{eq22});
			
			Constrain the boundary of individual $u_{i}$;
			
			Evaluate and record the fitness value of the $u_i$;
			
			$n\!f\!es$ = $n\!f\!es$ + 1;
						
			Select the off-spring individual $x_{i}$ by Eq.~(\ref{eq4});
		}
            
		Update external archive;
		
		Update $H_F$ and $H_{Cr}$ by Eq.~(\ref{eq14})-(\ref{eq17});
  
		Update sub-population racios $\gamma_1^{(k+1)}$ and $\gamma_2^{(k+1)}$ by Eq.~(\ref{eq6})-(\ref{eq8});
  
		Update the population size $N^{(k+1)}$ by Eq.~(\ref{eq21});
	}
\end{algorithm} 

\subsubsection{The pseudo-code of RDE}
The pseudo-code of RDE is given in Algorithm~\ref{RDE}.

\section{Experimental Results}
\subsection{Benchmark Functions}
The performance of the proposed algorithm RDE is tested on the CEC 2024 Competition for Single-Objective Real Parameter Numerical Optimization\cite{wu2017problem}. The benchmark suite, CEC2024, comprises 29 test functions (with $30D$) with diverse features, including unimodal functions ($f_1-f_2$), multimodal functions ($f3-f9$), hybrid funtions ($f10-f19$), and composition functions ($f20-f29$). The variables in hybrid functions are randomly divided into subcomponents, and different basic functions are used for each subcomponent. The composition functions utilize basic benchmark functions to create problems with a randomly located global optimum and several randomly located deep local optima, making them more challenging to solve.

\subsection{Parameter Settings}
In this study, the algorithm parameter settings are set as follows:
\begin{itemize}
    \item [1)] population maximum and minimum sizes $N_{min} = 4$ and $N_{max} = 18\cdot D$;
    \item [2)] success history archive memory size $H = 5$;
    \item [3)] rank greediness factor $k_r = 3$;
    \item [4)] external archive size $Ar = 1$;
    \item [5)] differential mutation $p$ value: $p_{max} = 0.25$;
    \item [6)] initialized sub-population resource ratios $\gamma_1 = \gamma_2 = 0.5$;
    \item [7)] initialized scale factor and crossover rate in memory archive $\mu_F = 0.3$ and $\mu_{Cr} = 0.8$;
    \item [8)] Cauthy pertubation strategy rate $p_r = 0.2$;
\end{itemize}

\begin{table*}[t!]
\centering
\caption{Experimental comparison results between RDE and other competitors on the 29 benchmark functions in CEC2024.} \label{ExperimentalResults}
\centering
\scriptsize
\renewcommand{\arraystretch}{0.9} 
\setlength{\tabcolsep}{3pt} 
\scalebox{1}{
\begin{tabular}{|c|cc|ccc|ccc|ccc|ccc|ccc|}
\hline
\multirow{2}{*}{Problem} & \multicolumn{2}{c|}{RDE}              & \multicolumn{3}{c|}{LSHADE-RSP}                    & \multicolumn{3}{c|}{iLSHADE-RSP}                   & \multicolumn{3}{c|}{HSES}                          & \multicolumn{3}{c|}{EBOwithCMAR}                   & \multicolumn{3}{c|}{LSHADE}                        \\ \cline{2-18} 
                         & Mean              & SD                & Mean              & SD                & W          & Mean              & SD                & W          & Mean              & SD                & W          & Mean              & SD                & W          & Mean              & SD                & W          \\ \hline
1                        & \textbf{0.00E+00} & \textbf{0.00E+00} & \textbf{0.00E+00} & \textbf{0.00E+00} & \textbf{=} & \textbf{0.00E+00} & \textbf{0.00E+00} & \textbf{=} & \textbf{0.00E+00} & \textbf{0.00E+00} & \textbf{=} & \textbf{0.00E+00} & \textbf{0.00E+00} & \textbf{=} & \textbf{0.00E+00} & \textbf{0.00E+00} & \textbf{=} \\
2                        & \textbf{0.00E+00} & \textbf{0.00E+00} & \textbf{0.00E+00} & \textbf{0.00E+00} & \textbf{=} & \textbf{0.00E+00} & \textbf{0.00E+00} & \textbf{=} & \textbf{0.00E+00} & \textbf{0.00E+00} & \textbf{=} & \textbf{0.00E+00} & \textbf{0.00E+00} & \textbf{=} & \textbf{0.00E+00} & \textbf{0.00E+00} & \textbf{=} \\
3                        & 2.16E+01          & 1.53E+00          & 5.86E+01          & 0.00E+00          & +          & 2.23E+01          & 8.92E-01          & +          & \textbf{2.66E+00} & \textbf{1.90E+00} & -          & 5.65E+01          & 1.11E+01          & +          & 5.86E+01          & 3.41E-14          & +          \\
4                        & 7.65E+00          & 1.85E+00          & 8.43E+00          & 3.48E+00          & =          & 6.89E+00          & 1.62E+00          & -          & 8.84E+00          & 3.76E+00          & =          & \textbf{2.78E+00} & \textbf{1.74E+00} & -          & 6.70E+00          & 1.40E+00          & -          \\
5                        & \textbf{0.00E+00} & \textbf{0.00E+00} & \textbf{0.00E+00} & \textbf{0.00E+00} & \textbf{=} & 1.34E-08          & 4.11E-08          & +          & \textbf{0.00E+00} & \textbf{0.00E+00} & \textbf{=} & \textbf{0.00E+00} & \textbf{0.00E+00} & \textbf{=} & 2.68E-09          & 1.92E-08          & =          \\
6                        & 3.86E+01          & 1.73E+00          & 4.12E+01          & 3.83E+00          & +          & 3.89E+01          & 1.69E+00          & =          & 3.91E+01          & 3.20E+00          & =          & \textbf{3.35E+01} & \textbf{8.37E-01} & \textbf{-} & 3.73E+01          & 1.37E+00          & -          \\
7                        & 8.37E+00          & 1.92E+00          & 9.29E+00          & 4.36E+00          & =          & 7.74E+00          & 1.77E+00          & =          & 7.65E+00          & 2.74E+00          & -          & \textbf{2.02E+00} & \textbf{1.32E+00} & \textbf{-} & 6.94E+00          & 1.75E+00          & -          \\
8                        & \textbf{0.00E+00} & \textbf{0.00E+00} & \textbf{0.00E+00} & \textbf{0.00E+00} & \textbf{=} & \textbf{0.00E+00} & \textbf{0.00E+00} & \textbf{=} & \textbf{0.00E+00} & \textbf{0.00E+00} & \textbf{=} & \textbf{0.00E+00} & \textbf{0.00E+00} & \textbf{=} & \textbf{0.00E+00} & \textbf{0.00E+00} & \textbf{=} \\
9                        & 1.43E+03          & 2.43E+02          & 2.87E+03          & 8.64E+02          & +          & 1.68E+03          & 2.70E+02          & +          & \textbf{9.63E+02} & \textbf{3.37E+02} & \textbf{-} & 1.41E+03          & 2.15E+02          & =          & 1.41E+03          & 2.35E+02          & =          \\
10                       & 3.93E+00          & 3.38E+00          & \textbf{3.13E+00} & \textbf{8.50E+00} & \textbf{-} & 3.30E+00          & 5.56E+00          & =          & 1.79E+01          & 2.59E+01          & +          & 4.49E+00          & 8.77E+00          & =          & 3.20E+01          & 2.89E+01          & +          \\
11                       & 2.50E+02          & 1.66E+02          & 9.66E+01          & 7.48E+01          & -          & 1.19E+02          & 8.37E+01          & -          & \textbf{3.84E+01} & \textbf{1.05E+02} & \textbf{-} & 4.63E+02          & 2.63E+02          & +          & 1.13E+03          & 3.17E+02          & +          \\
12                       & \textbf{1.44E+01} & \textbf{5.80E+00} & 1.71E+01          & 3.61E+00          & +          & 1.81E+01          & 4.54E+00          & +          & 2.77E+01          & 1.29E+01          & +          & 1.49E+01          & 6.25E+00          & =          & 1.51E+01          & 5.46E+00          & =          \\
13                       & 2.02E+01          & 5.98E+00          & 2.15E+01          & 1.15E+00          & =          & 2.16E+01          & 1.14E+00          & =          & \textbf{1.35E+01} & \textbf{9.87E+00} & \textbf{-} & 2.19E+01          & 3.84E+00          & +          & 2.07E+01          & 4.89E+00          & =          \\
14                       & 1.46E+00          & 8.86E-01          & \textbf{8.25E-01} & \textbf{5.36E-01} & -          & 1.05E+00          & 6.39E-01          & =          & 4.72E+00          & 4.74E+00          & +          & 3.69E+00          & 2.15E+00          & +          & 3.05E+00          & 1.39E+00          & +          \\
15                       & 2.72E+01          & 4.49E+01          & \textbf{1.44E+01} & \textbf{5.46E+00} & \textbf{=} & 1.65E+01          & 5.25E+00          & +          & 2.51E+02          & 2.16E+02          & +          & 4.26E+01          & 5.69E+01          & +          & 4.87E+01          & 4.35E+01          & +          \\
16                       & 2.94E+01          & 1.10E+01          & 3.80E+01          & 9.81E+00          & +          & 3.55E+01          & 5.04E+00          & +          & \textbf{2.46E+01} & \textbf{3.53E+01} & \textbf{-} & 2.98E+01          & 7.50E+00          & =          & 3.32E+01          & 5.36E+00          & =          \\
17                       & 2.03E+01          & 2.86E+00          & 2.08E+01          & 2.32E-01          & +          & 2.00E+01          & 3.88E+00          & +          & \textbf{1.96E+01} & \textbf{4.96E+00} & \textbf{+} & 2.21E+01          & 1.09E+00          & +          & 2.19E+01          & 1.15E+00          & +          \\
18                       & 3.35E+00          & 9.36E-01          & \textbf{3.21E+00} & \textbf{7.47E-01} & =          & 3.34E+00          & 7.47E-01          & =          & 4.07E+00          & 2.00E+00          & =          & 8.04E+00          & 2.28E+00          & +          & 4.92E+00          & 1.61E+00          & +          \\
19                       & \textbf{2.57E+01} & \textbf{6.66E+00} & 2.65E+01          & 7.16E+00          & =          & 3.22E+01          & 5.81E+00          & +          & 1.44E+02          & 3.17E+01          & +          & 3.57E+01          & 7.50E+00          & +          & 3.20E+01          & 5.97E+00          & +          \\
20                       & 2.08E+02          & 2.05E+00          & 2.09E+02          & 4.07E+00          & +          & 2.08E+02          & 1.97E+00          & =          & 2.08E+02          & 3.74E+00          & =          & \textbf{1.99E+02} & \textbf{2.02E+01} & \textbf{-} & 2.07E+02          & 1.30E+00          & =          \\
21                       & \textbf{1.00E+02} & \textbf{0.00E+00} & \textbf{1.00E+02} & \textbf{0.00E+00} & \textbf{=} & \textbf{1.00E+02} & \textbf{0.00E+00} & \textbf{=} & \textbf{1.00E+02} & \textbf{0.00E+00} & \textbf{=} & \textbf{1.00E+02} & \textbf{0.00E+00} & \textbf{=} & 1.00E+02          & 1.00E-13          & +          \\
22                       & \textbf{3.46E+02} & \textbf{3.38E+00} & 3.52E+02          & 4.14E+00          & +          & 3.51E+02          & 3.04E+00          & +          & 3.52E+02          & 8.59E+00          & +          & 3.51E+02          & 3.51E+00          & +          & 3.50E+02          & 2.92E+00          & +          \\
23                       & 4.23E+02          & 2.38E+00          & 4.27E+02          & 2.68E+00          & +          & 4.26E+02          & 1.87E+00          & +          & 4.21E+02          & 3.39E+00          & -          & \textbf{4.18E+02} & \textbf{4.55E+01} & \textbf{+} & 4.26E+02          & 1.70E+00          & +          \\
24                       & \textbf{3.79E+02} & \textbf{1.56E-01} & 3.87E+02          & 8.02E-03          & +          & 3.79E+02          & 2.29E-01          & +          & 3.87E+02          & 2.65E-02          & +          & 3.87E+02          & 7.56E-01          & +          & 3.87E+02          & 2.28E-02          & +          \\
25                       & 8.94E+02          & 4.27E+01          & 9.01E+02          & 4.03E+01          & =          & 9.24E+02          & 3.89E+01          & +          & 8.77E+02          & 2.00E+02          & =          & \textbf{5.37E+02} & \textbf{3.06E+02} & \textbf{-} & 9.17E+02          & 3.30E+01          & +          \\
26                       & \textbf{4.73E+02} & \textbf{6.62E+00} & 4.96E+02          & 8.00E+00          & +          & 4.77E+02          & 6.21E+00          & +          & 5.20E+02          & 9.26E+00          & +          & 5.02E+02          & 4.03E+00          & +          & 5.03E+02          & 6.46E+00          & +          \\
27                       & 3.18E+02          & 4.14E+01          & \textbf{3.02E+02} & \textbf{1.60E+01} & \textbf{-} & 3.04E+02          & 2.23E+01          & -          & 3.18E+02          & 3.98E+01          & =          & 3.08E+02          & 2.88E+01          & =          & 3.22E+02          & 4.63E+01          & +          \\
28                       & \textbf{3.88E+02} & \textbf{2.66E+01} & 4.38E+02          & 1.84E+01          & +          & 3.92E+02          & 2.08E+01          & +          & 4.55E+02          & 5.40E+01          & +          & 4.33E+02          & 1.13E+01          & +          & 4.32E+02          & 6.13E+00          & +          \\
29                       & \textbf{8.48E+02} & \textbf{3.99E+02} & 1.97E+03          & 1.04E+01          & +          & 1.01E+03          & 3.35E+02          & +          & 2.06E+03          & 5.12E+01          & +          & 1.99E+03          & 4.21E+01          & +          & 1.98E+03          & 3.79E+01          & +          \\ \hline
W/T/L                    & \multicolumn{2}{c|}{$-$/$-$/$-$}                 & \multicolumn{3}{c|}{13/13/4}                       & \multicolumn{3}{c|}{15/12/3}                       & \multicolumn{3}{c|}{11/12/7}                       & \multicolumn{3}{c|}{14/11/5}                       & \multicolumn{3}{c|}{17/10/3}                       \\ \hline
\end{tabular}
}
\end{table*}

\subsection{Experimental Settings}
According to the guidelines of CEC2024, 25 independent runs are implemented. The maximum evaluation number is set to $max\_n\!f\!es = 10000\cdot D$, where the $D$ is the dimensionality of the optimization problems. The search range is set to $[-100, 100]^D$. Fitness errors
less than $1\times 10^{−8}$ are considered to 0. The experiments were performed using the following system: CPU: Intel\textsuperscript{\textregistered} Core\texttrademark i7-12700kf 3.6GHz, 32GB DDR4-3600MHz RAM, Programming Language: C++, and Operator System: Windows 10.

\begin{table}[t!]
\centering
\scriptsize
\renewcommand{\arraystretch}{1.1} 
\setlength{\tabcolsep}{4pt} 
\caption{Algorithm Complexity of RDE}
\label{Algorithm_Complexity}
\begin{tabular}{|c|c|c|c|c|}
\hline
 & $T_0$ & $T_1$ & $\hat{T}_2$ & $(\hat{T}_2 - T_1)/T_0$ \\ \hline
D = 30  & 27    & 351   & 484.6       & 4.95                    \\ \hline
\end{tabular}
\end{table}

\subsection{Statistical Results}
The Wilcoxon rank-sum test is utilized to determine if there is a statistically significant difference between two algorithms in solving a problem, with a significance level of $\alpha = 0.05$. Each problem is tested 51 times independently in a row. The symbol ``$+$" denotes the superior algorithm when there is a significant difference between the pair, while ``$-$" indicates the relatively inferior one. ``$=$" signifies that no significant difference exists between the pair of algorithms.

The performance of the RDE was compared to the other compatitors which showed excellent performance on previous CEC benchmark suites, including:
\begin{itemize}
    \item [1)] LSHADE-RSP\cite{stanovov2018lshade}: LSHADE Algorithm with Rank-Based Selective Pressure Strategy for Solving CEC 2017 Benchmark Problems;
    \item [2)] iLSHADE-RSP\cite{choi2021improved}: An improved LSHADE-RSP algorithm with the Cauchy perturbation;
    \item [3)] HSES\cite{zhang2018hybrid}: Hybrid Sampling Evolution Strategy for Solving Single Objective Bound Constrained Problems;
    \item [4)] EBOwithCMAR\cite{kumar2017improving}: Effective Butterfly Optimizer using Covariance Matrix Adapted Retreat phase;
    \item [5)] LSHADE\cite{tanabe2014improving}: Improving the search performance of SHADE using linear population size reduction.
\end{itemize}

As shown in Table~\ref{ExperimentalResults}, RDE demonstrates superiority over competitive algorithms on CEC2024. In addition, RDE also performs well on relatively more complex composition functions ($f20-f29$), indicating the effectiveness of recombination research. Compared with LSHADE-RSP, iLSHADE-RSP, HSES, EBOwithCMAR and LSHADE, RDE achieves better solutions on 13, 15, 11, 14 and 17 problems respectively and fails to beat opponents on 4, 3, 7, 5 and 3 problems respectively. Although RDE shows significant advantages over advanced variants of the DE series, the performance gap is not significantly widened compared to HSES.

\subsection{Algorithm Complexity}
Table~\ref{Algorithm_Complexity} shows that the algorithm complexity of the proposed RDE is 4.95. $T_0$ is the time result by running the assessing test program in \cite{wu2017problem}. $T_1$ is the time cost result by running the 18th function and $T_2$ is the time cost result of averaging 5 runs of the algorithm on $f_{18}$, respectively. Both of them are with 200000 evaluations.

\section{Conclusion}
In this study, we conducted a combined research on recent advances in advanced DE variants and proposed an effective combination improvement scheme to further promote the performance of DE in complex optimization problems. On the CEC2024 benchmark suite, RDE showed significant performance advantages over some advanced variants of DE. Although the ES variant HSES achieved similar performance in 2018, due to the simplicity, efficiency and stable performance of DE, it seems to have received more attention and development from researchers. Although we also attempted to combine other recent DE improvement strategies besides those shown in this paper, no significant results were obtained. Therefore, various sub-strategies mentioned in this paper may be worth prioritizing for future development of DE. It is worth noting that compared with algorithms such as iLSHADE-RSP, RDE adopts fewer strategies but has better performance.

\section*{Acknowledgment}
This research was partially supported by the Japan Society for the Promotion of Science (JSPS) KAKENHI under Grant JP22H03643, Japan Science and Technology Agency (JST) Support for Pioneering Research Initiated by the Next Generation (SPRING) under Grant JPMJSP2145, and JST through the Establishment of University Fellowships towards the Creation of Science Technology Innovation under Grant JPMJFS2115.

\bibliographystyle{IEEEtran}
\bibliography{RDE}

\begin{thebibliography}{10}
\providecommand{\url}[1]{#1}
\csname url@samestyle\endcsname
\providecommand{\newblock}{\relax}
\providecommand{\bibinfo}[2]{#2}
\providecommand{\BIBentrySTDinterwordspacing}{\spaceskip=0pt\relax}
\providecommand{\BIBentryALTinterwordstretchfactor}{4}
\providecommand{\BIBentryALTinterwordspacing}{\spaceskip=\fontdimen2\font plus
\BIBentryALTinterwordstretchfactor\fontdimen3\font minus
  \fontdimen4\font\relax}
\providecommand{\BIBforeignlanguage}[2]{{%
\expandafter\ifx\csname l@#1\endcsname\relax
\typeout{** WARNING: IEEEtran.bst: No hyphenation pattern has been}%
\typeout{** loaded for the language `#1'. Using the pattern for}%
\typeout{** the default language instead.}%
\else
\language=\csname l@#1\endcsname
\fi
#2}}
\providecommand{\BIBdecl}{\relax}
\BIBdecl

\bibitem{gao2016ant}
S.~Gao, Y.~Wang, J.~Cheng, Y.~Inazumi, and Z.~Tang, ``Ant colony optimization
  with clustering for solving the dynamic location routing problem,''
  \emph{Applied Mathematics and Computation}, vol. 285, pp. 149--173, 2016.

\bibitem{gao2021state}
S.~Gao, K.~Wang, S.~Tao, T.~Jin, H.~Dai, and J.~Cheng, ``A state-of-the-art
  differential evolution algorithm for parameter estimation of solar
  photovoltaic models,'' \emph{Energy {C}onversion and {M}anagement}, vol. 230,
  p. 113784, 2021.

\bibitem{8409490}
S.~Gao, M.~Zhou, Y.~Wang, J.~Cheng, H.~Yachi, and J.~Wang, ``Dendritic neuron
  model with effective learning algorithms for classification, approximation,
  and prediction,'' \emph{IEEE Transactions on Neural Networks and Learning
  Systems}, vol.~30, no.~2, pp. 601--614, 2019.

\bibitem{8937719}
S.~Gao, Y.~Yu, Y.~Wang, J.~Wang, J.~Cheng, and M.~Zhou, ``Chaotic local
  search-based differential evolution algorithms for optimization,'' \emph{IEEE
  Transactions on Systems, Man, and Cybernetics: Systems}, vol.~51, no.~6, pp.
  3954--3967, 2021.

\bibitem{wang2023spherical}
K.~Wang, Y.~Wang, S.~Tao, Z.~Cai, Z.~Lei, and S.~Gao, ``Spherical search
  algorithm with adaptive population control for global continuous optimization
  problems,'' \emph{Applied Soft Computing}, vol. 132, p. 109845, 2023.

\bibitem{sohail2023genetic}
A.~Sohail, ``Genetic algorithms in the fields of artificial intelligence and
  data sciences,'' \emph{Annals of Data Science}, vol.~10, no.~4, pp.
  1007--1018, 2023.

\bibitem{yarat2021comparative}
S.~Yarat, S.~Senan, and Z.~Orman, ``A comparative study on pso with other
  metaheuristic methods,'' \emph{Applying Particle Swarm Optimization: New
  Solutions and Cases for Optimized Portfolios}, pp. 49--72, 2021.

\bibitem{ahmad2022differential}
M.~F. Ahmad, N.~A.~M. Isa, W.~H. Lim, and K.~M. Ang, ``Differential evolution:
  A recent review based on state-of-the-art works,'' \emph{Alexandria
  Engineering Journal}, vol.~61, no.~5, pp. 3831--3872, 2022.

\bibitem{tanabe2013success}
R.~Tanabe and A.~Fukunaga, ``Success-history based parameter adaptation for
  differential evolution,'' in \emph{2013 IEEE Congress on Evolutionary
  Computation}.\hskip 1em plus 0.5em minus 0.4em\relax IEEE, 2013, pp. 71--78.

\bibitem{tanabe2014improving}
R.~Tanabe and A.~S. Fukunaga, ``Improving the search performance of {SHADE}
  using linear population size reduction,'' in \emph{2014 {IEEE} {C}ongress on
  {E}volutionary {C}omputation ({CEC})}.\hskip 1em plus 0.5em minus 0.4em\relax
  IEEE, 2014, pp. 1658--1665.

\bibitem{stanovov2018lshade}
V.~Stanovov, S.~Akhmedova, and E.~Semenkin, ``{LSHADE} algorithm with
  rank-based selective pressure strategy for solving {CEC} 2017 benchmark
  problems,'' in \emph{2018 {IEEE} {C}ongress on {E}volutionary {C}omputation
  ({CEC})}.\hskip 1em plus 0.5em minus 0.4em\relax IEEE, 2018, pp. 1--8.

\bibitem{choi2021improved}
T.~J. Choi and C.~W. Ahn, ``An improved lshade-rsp algorithm with the cauchy
  perturbation: ilshade-rsp,'' \emph{Knowledge-Based Systems}, vol. 215, p.
  106628, 2021.

\bibitem{mohamed2019novel}
A.~W. Mohamed, A.~A. Hadi, and K.~M. Jambi, ``Novel mutation strategy for
  enhancing shade and lshade algorithms for global numerical optimization,''
  \emph{Swarm and Evolutionary Computation}, vol.~50, p. 100455, 2019.

\bibitem{storn1997differential}
R.~Storn and K.~Price, ``Differential evolution--a simple and efficient
  heuristic for global optimization over continuous spaces,'' \emph{Journal of
  Global Optimization}, vol.~11, pp. 341--359, 1997.

\bibitem{zhang2009jade}
J.~Zhang and A.~C. Sanderson, ``Jade: adaptive differential evolution with
  optional external archive,'' \emph{IEEE Transactions on Evolutionary
  Computation}, vol.~13, no.~5, pp. 945--958, 2009.

\bibitem{holland1992genetic}
J.~H. Holland, ``Genetic algorithms,'' \emph{Scientific American}, vol. 267,
  no.~1, pp. 66--73, 1992.

\bibitem{jebari2013selection}
K.~Jebari, M.~Madiafi \emph{et~al.}, ``Selection methods for genetic
  algorithms,'' \emph{International Journal of Emerging Sciences}, vol.~3,
  no.~4, pp. 333--344, 2013.

\bibitem{brest2017single}
J.~Brest, M.~S. Mau{\v{c}}ec, and B.~Bo{\v{s}}kovi{\'c}, ``Single objective
  real-parameter optimization: algorithm j{SO},'' in \emph{2017 {IEEE} Congress
  on {E}volutionary {C}omputation ({CEC})}.\hskip 1em plus 0.5em minus
  0.4em\relax IEEE, 2017, pp. 1311--1318.

\bibitem{brest2016shade}
J.~Brest, M.~S. Mau\v{c}ec, and B.~Bo{\v{s}}kovi{\'c}, ``il-shade: Improved
  l-shade algorithm for single objective real-parameter optimization,'' in
  \emph{2016 IEEE Congress on Evolutionary Computation (CEC)}.\hskip 1em plus
  0.5em minus 0.4em\relax IEEE, 2016, pp. 1188--1195.

\bibitem{wu2017problem}
G.~Wu, R.~Mallipeddi, and P.~N. Suganthan, ``Problem definitions and evaluation
  criteria for the cec 2017 competition on constrained real-parameter
  optimization,'' \emph{National University of Defense Technology, Changsha,
  Hunan, PR China and Kyungpook National University, Daegu, South Korea and
  Nanyang Technological University, Singapore, Technical Report}, 2017.

\bibitem{zhang2018hybrid}
G.~Zhang and Y.~Shi, ``Hybrid sampling evolution strategy for solving single
  objective bound constrained problems,'' in \emph{2018 IEEE Congress on
  Evolutionary Computation (CEC)}.\hskip 1em plus 0.5em minus 0.4em\relax IEEE,
  2018, pp. 1--7.

\bibitem{kumar2017improving}
A.~Kumar, R.~K. Misra, and D.~Singh, ``Improving the local search capability of
  effective butterfly optimizer using covariance matrix adapted retreat
  phase,'' in \emph{2017 IEEE Congress on Evolutionary Computation
  (CEC)}.\hskip 1em plus 0.5em minus 0.4em\relax IEEE, 2017, pp. 1835--1842.

\end{thebibliography}
\end{CJK}
\end{document}